\documentclass[a4paper]{article}%
\usepackage{amsmath}%
\usepackage{amsfonts}%
\usepackage{amssymb}%
\usepackage{graphicx}
\usepackage{fullpage}

\usepackage{math_defs}

\begin{document}

\title{Bayesian SPLDA}
\author{Jes\'{u}s Villalba\\\\
  Communications Technology Group (GTC),\\ Aragon Institute
  for Engineering Research (I3A),\\ University of Zaragoza, Spain\\
  \small \tt villalba@unizar.es}
\date{May 15, 2012}
\maketitle

\section{Introduction}

In this document we are going to derive the equations needed to
implement a Variational Bayes estimation of the parameters of the
SPLDA model~\cite{villalba-splda}. 
This can be used to adapt the SPLDA from one database to
another with few development data or to implement the fully Bayesian
recipe~\cite{villalba-bay2cov-is2011}.
Our approach is similar to Bishop's VB PPCA
in~\cite{Bishop1999}.

\section{The Model}

\subsection{SPLDA}

\begin{figure}[th]
  \begin{center}
    \includegraphics[width=0.60\textwidth]
    {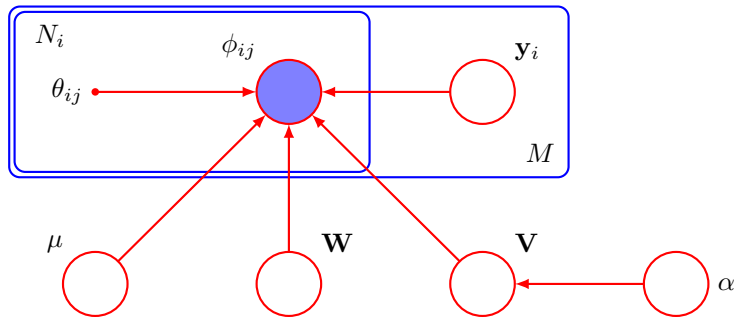}
  \end{center}
  \vspace{-0.5cm}
  \caption{BN for Bayesian SPLDA model.}
  \label{fig:bn_baysplda}
\end{figure}

SPLDA is a linear generative model represented in
Figure~\ref{fig:bn_baysplda}.

An
i-vector $\phivec$ of speaker $i$ can be written as:
\begin{equation}
  \phivec_{ij}=\muvec+\Vmat\yvec_{i}+\epsilon_{ij}
  \label{eq:baysplda_model}
\end{equation} 
where $\muvec$ is a speaker independent mean, $\Vmat$ is the eigenvoices
matrix, $\yvec_i$ is the speaker factor vector, and $\epsilon$ is a channel
offset. 

We assume the following priors for the variables:
\begin{align}
  \label{eq:baysplda_yprior}
  \yvec_i&\sim\Gauss{\yvec_i}{\zerovec}{\Imat} \\
  \label{eq:baysplda_eprior}
  \epsilon_{ij}&\sim\Gauss{\epsilon_{ij}}{\zerovec}{\iWmat}
\end{align}
where $\mathcal{N}$ denotes a Gaussian distribution; $\Wmat$
is the within class precision matrix.

\subsection{Notation}

We are going to introduce some notation:
\begin{itemize}
\item Let $\Phimatd$ be the development i-vectors dataset.
\item Let $\Phimatt=\left\{l,r\right\}$ be the test i-vectors.
\item Let $\Phimat$ be any of the previous datasets.
\item Let $\thetad$ be the labelling of the development dataset. It
  partitions the $N_d$ i-vectors into $M_d$ speakers.
\item Let $\thetat$ be the labelling of the test set, so that $\thetat
  \in \left\{\tar,\nontar\right\}$, where $\tar$ is the hypothesis
  that $l$ and $r$ belong to the same speaker and $\nontar$ is the
  hypothesis that they belong to different speakers.
\item Let $\theta$ be any of the previous labellings.
\item Let $\spk_i$ be the i-vectors belonging to the speaker $i$.
\item Let $\phivec_{ij}$ the i-vector $j$ of speaker $i$.  
\item Let $\Ymatd$ be the speaker identity variables of the
  development set. We will have as many identity variables as speakers.
\item Let $\Ymatt$ be the speaker identity variables of the test set.
\item Let $\Ymat$ be any of the previous speaker identity variables
  sets.
\item Let $d$ be the i-vector dimension.
\item Let $n_y$ be the speaker factor dimension.
\item Let $\model=\left(\muvec,\Vmat,\Wmat\right)$ be the set of all
  the parameters.
\end{itemize}

\section{Sufficient statistics}

We define the sufficient statistics for speaker $i$. The zero-order
statistic is the number of observations of speaker $i$
$N_i$. The first-order and second-order statistics are
\begin{align}
  \Fvec_i=&\sumjNi \phivec_{ij}\\
  \Smat_i=&\sumjNi \scatt{\phivec_{ij}}
\end{align} 
We define the centered statistics as
\begin{align}
  \Fbar_i=&\Fvec_i-N_i \muvec\\
  \Sbarmat_i=&\sumjNi \scattp{\phivec_{ij}-\muvec}
  =\Smat_i-\muvec\Fvec_i^T-\Fvec_i\muvec^T+N_i\scatt{\muvec}
\end{align}

We define the global statistics 
\begin{align}
  N&=\sumiM N_i \\
  \Fvec&=\sumiM \Fvec_i \\
  \Fbar&=\sumiM \Fbar_i \\
  \Smat&=\sumiM \Smat_i\\
  \Sbarmat&=\sumiM \Sbarmat_i
\end{align}

\section{Data conditional likelihood}

The likelihood of the data given the hidden variables for speaker $i$
is 
\begin{align}
  \label{eq:baysplda_cond1}
  \lnProb{\spk_i|\yvec_i,\muvec,\Vmat,\Wmat}=&\sumjNi 
  \ln\Gauss{\phivec_{ij}}{\muvec+\Vmat\yvec_i}{\iWmat} \\
  \label{eq:baysplda_cond2}
  =&\frac{N_{i}}{2}\lndet{\frac{\Wmat}{2\pi}}
  -\med\sumjNi \mahP{\phivec_{ij}}{\muvec-\Vmat\yvec_i}{\Wmat}\\
  \label{eq:baysplda_cond3}
  =&\frac{N_{i}}{2}\lndet{\frac{\Wmat}{2\pi}}
  -\med\trace\left(\Wmat\Sbarmat_{i}\right)
  +\yvec_i^{T}\Vmat^{T}\Wmat\Fbar_{i}
  -\frac{N_{i}}{2}\yvec_i^T\Vmat^T\Wmat\Vmat\yvec_i
\end{align}

We can write this likelihood in another form:

\begin{align}
  \label{eq:baysplda_cond4}
  \lnProb{\spk_i|\yvec_i,\muvec,\Vmat,\Wmat}=&
  \frac{N_{i}}{2}\lndet{\frac{\Wmat}{2\pi}}
  -\med\trace\left(\Wmat\left(\Smat_{i}
      -2\Fvec_{i}\muvec^{T}
      +N_{i}\scatt{\muvec}
    \right.\right.\\
  &\left.\left.-2 \left(\Fvec_{i}-N_{i}\muvec\right)\yvec_i^{T}\Vmat^{T}
      +N_{i} \Vmat\yvec_i\yvec_i^T\Vmat^T\right)\right)
\end{align}

We can write this likelihood in another form if we define:
\begin{align}
  \ytildevec_i=
  \begin{bmatrix}
    \yvec_i\\
    1
  \end{bmatrix}
  , & \quad \Vtildemat=
  \begin{bmatrix}
    \Vmat & \muvec
  \end{bmatrix}
\end{align}
Then
\begin{align}
  \label{eq:baysplda_cond5}
  \lnProb{\spk_i|\yvec_i,\muvec,\Vmat,\Wmat}=&\sumjNi \ln
  \Gauss{\phivec_{ij}}{\Vtildemat\ytildevec_i}{\iWmat} \\
  \label{eq:baysplda_cond6}
  =&\frac{N_i}{2}\lndet{\frac{\Wmat}{2\pi}}
  -\med\sum_{j=1}^{N_i}\mahP{\phivec_{ij}}{\Vtildemat\ytildevec_i}{\Wmat}\\
  \label{eq:baysplda_cond7}
  =&\frac{N_i}{2}\lndet{\frac{\Wmat}{2\pi}}-\med\trace\left(\Wmat\Smat_i\right)
  +\ytildevec_i^{T}\Vtildemat^{T}\Wmat\Fvec_i
  -\frac{N_i}{2}\ytildevec_i^T\Vtildemat^T\Wmat\Vtildemat\ytildevec_i\\
  \label{eq:baysplda_cond8}
  =&\frac{N_i}{2}\lndet{\frac{\Wmat}{2\pi}}
  -\med\trace\left(\Wmat\left(\Smat_i-2\Fvec_i\ytildevec_i^{T}\Vtildemat^{T}
      +N_i \Vtildemat\ytildevec_i\ytildevec_i^T\Vtildemat^T\right)\right)
\end{align}

\section{Variational inference with Gaussian-Gamma priors for $\Vmat$,
  Gaussian for $\muvec$ and Wishart for $\Wmat$ (informative and non-informative)}
\label{sec:baysplda_v1}

\subsection{Model priors}
\label{sec:baysplda_v1_priors}

We introduce a \emph{hierarchical} prior $\Prob{\Vmat|\alphavec}$ over
the matrix $\Vmat$ governed by a $n_y$ dimensional vector of
hyperparameters where $n_y$ is the dimension of the factors. Each
hyperparameter controls one of the columns of the matrix $\Vmat$
through a conditional Gaussian distribution of the form:
\begin{align}
  \Prob{\Vmat|\alphavec}=
  \prod_{q=1}^{n_y}\left(\frac{\alpha_q}{2\pi}\right)^{d/2}
  \exp\left(-\med\alpha_q\vvec_q^T\vvec_q\right)
\end{align}
where $\vvec_q$ are the columns of $\Vmat$. Each $\alpha_q$ controls
the inverse variance of the corresponding $\vvec_q$. If a particular
$\alpha_q$ has a posterior distribution concentrated at large
values, the corresponding $\vvec_q$ will tend to be small, and that
direction of the latent space will be effectively 'switched off'. 

We define a prior for $\alphavec$:
\begin{align}
  \Prob{\alphavec}=
  \prod_{q=1}^{n_y}\Gammad{\alpha_q}{a_{\alpha}}{b_{\alpha}}
\end{align}
where $\mathcal{G}$ denotes the Gamma distribution. Bishop defines broad
priors setting $a=b=10^{-3}$. 

We place a Gaussian prior for the mean $\muvec$:
\begin{align}
  \Prob{\muvec}=\Gauss{\muvec}{\muvec_0}{\diag(\betavec)^{-1}}
\end{align}
We will consider the case where each dimension has
different precision and the case with isotropic precision 
($\diag(\betavec)=\beta\Imat$). 

Finally, we put a Wishart prior on $\Wmat$,
\begin{align}
  \Prob{\Wmat}=\Wishart{\Wmat}{\Psimat_{0}}{\nud}
\end{align}
We can make the Wishart prior non-informative like
in~\cite{villalba-bay2cov}. 
\begin{align}
  \label{eq:jpriorb1}
  \Prob{\Wmat}&=\lim_{k \to 0}\Wishart{\Wmat}{\Wmat_0/k}{k}\\
  \label{eq:jpriorb2}
  &=\alpha \left|\Wmat\right|^{-(d+1)/2}
\end{align}

\subsection{Variational distributions}
\label{sec:baysplda_v1_q}

We write the joint distribution of the observed and latent variables:
\begin{align}
  \Prob{\Phimat,\Ymat,\muvec,\Vmat,\Wmat,\alphavec
    |\muvec_0,\betavec,a_{\alpha},b_{\alpha}}=
  \Prob{\Phimat|\Ymat,\muvec,\Vmat,\Wmat}\Prob{\Ymat}
  \Prob{\Vmat|\alphavec}\Prob{\alphavec|a,b}
  \Prob{\muvec|\muvec_0,\beta}\Prob{\Wmat}
\end{align}
Following, the conditioning on
$\left(\muvec_0,\betavec,a_{\alpha},b_{\alpha}\right)$ 
will be dropped for convenience. 

Now, we consider the partition of the posterior:
\begin{align}
  \Prob{\Ymat,\muvec,\Vmat,\Wmat,\alphavec|\Phimat}\approx
  \q{\Ymat,\muvec,\Vmat,\Wmat,\alphavec}=
  \q{\Ymat}\prod_{r=1}^d\q{\vtildevec'_r}\q{\Wmat}\q{\alphavec}
\end{align}
where $\vtildevec'_r$ is a column vector containing the $r^{th}$ row of
$\Vtildemat$. If $\Wmat$ were a diagonal matrix the factorization 
$\prod_{r=1}^d\q{\vtildevec'_r}$ is not necessary because it arises
naturally when solving the posterior. However, for full covariance
$\Wmat$, the posterior of $\vec(\Vtildemat)$ is a Gaussian
with a huge full covariance matrix. Therefore, we force the factorization
to made the problem tractable. 

The optimum for $\qopt{\Ymat}$:
\begin{align}
  \lnqopt{\Ymat}=&
  \Expcond{\lnProb{\Phimat,\Ymat,\muvec,\Vmat,\Wmat,\alphavec}}
  {\muvec,\Vmat,\Wmat,\alphavec}+\const\\
  =&\Expcond{\lnProb{\Phimat|\Ymat,\muvec,\Vmat,\Wmat}}{\muvec,\Vmat,\Wmat}
  +\lnProb{\Ymat}+\const\\
  =&\sumiM \yvec_i^T\Exp{\Vmat^T\Wmat\left(\Fvec_i-N_i\muvec\right)} 
  -\med \yvec_i^T\left(\Imat+N_i\Exp{\Vmat^T\Wmat\Vmat}\right)\yvec_i+\const
\end{align}
Therefore $\qopt{\Ymat}$ is a product of Gaussian distributions.
\begin{align}
  \qopt{\Ymat}=&\prodiM \Gauss{\yvec_i}{\ybarvec_i}{\iLmatyi}\\
  \Lmatyi=&\Imat+N_i\Exp{\Vmat^T\Wmat\Vmat}\\
  \ybarvec_i=&\iLmatyi\Exp{\Vmat^T\Wmat\left(\Fvec_i-N_i\muvec\right)}\\
  =&\iLmatyi\left(\Exp{\Vmat}^T\Exp{\Wmat}\Fvec_i-N_i\Exp{\Vmat^T\Wmat\muvec}\right)
\end{align}

The optimum for $\qopt{\vtildevec'_r}$:
\begin{align}
  \lnqopt{\vtildevec'_r}=&
  \Expcond{\lnProb{\Phimat,\Ymat,\muvec,\Vmat,\Wmat,\alphavec}}
  {\Ymat,\Wmat,\alphavec,\vtildevec'_{s\neq r}}+\const\\
  =&\Expcond{\lnProb{\Phimat|\Ymat,\muvec,\Vmat,\Wmat}}
  {\Ymat,\Wmat,\vtildevec'_{s\neq r}}
  +\Expcond{\lnProb{\Vmat|\alphavec}}{\alphavec,\vvec'_{s\neq r}}
  +\Expcond{\lnProb{\muvec}}{\muvec_{s\neq r}}+\const\\
  =&-\med \sumiM \trace\left(\Exp{\Wmat}
    \left(-2\Fvec_i\Exp{\ytildevec_i}^{T}
      \Expcond{\Vtildemat}{\vtildevec'_{s\neq r}}^{T}
      +N_i 
      \Expcond{\Vtildemat\Exp{\ytildevec_i\ytildevec_i^T}\Vtildemat^T}
      {\vtildevec'_{s\neq r}}
    \right)\right) \nonumber\\
  &-\med \sum_{q=1}^{n_y} \Exp{\alpha_q}
  \Expcond{\vvec_q^T\vvec_q}{\vvec'_{s\neq r}}
  -\med \beta_{r}\left(\mu_r-\mu_{0_r}\right)^2
  +\const\\
  =&-\med\trace\left(\Exp{\Wmat}
    \left(-2\Cmat\Expcond{\Vtildemat}{\vtildevec'_{s\neq r}}^{T}
      +\Expcond{\Vtildemat\Rmatytilde\Vtildemat^T}
      {\vtildevec'_{s\neq r}}
    \right)\right) \nonumber\\
  &-\med \sum_{q=1}^{n_y} \Exp{\alpha_q}
  \Expcond{\vvec_q^T\vvec_q}{\vvec'_{s\neq r}}
  -\med \beta_{r}\left(\mu_r-\mu_{0_r}\right)^2
  +\const\\
  =&-\med\trace\left(-2\Expcond{\Vtildemat}
    {\vtildevec'_{s\neq r}}^{T}
    \Exp{\Wmat}\Cmat
    +\Expcond{\Vtildemat^{T}\Exp{\Wmat}\Vtildemat}
    {\vtildevec'_{s\neq r}}
    \Rmatytilde
  \right) \nonumber\\
  &  -\med \vvec_{r}'^T\diag\left(\Exp{\alphavec}\right)\vvec_{r}'
  -\med \beta_{r}\left(\mu_r-\mu_{0_r}\right)^2
  +\const\\
  =&-\med\trace\left(-2\sumsd\vtildevec'_{r}\wbar_{rs}\Cmat_{s}
    +2\sum_{s\neq r}
    \vtildevec'_{r}\wbar_{rs}\Exp{\vtildevec'_{s}}^{T}\Rmatytilde
    +\vtildevec'_r\wbar_{rr}\vtildevec_r'^{T}\Rmatytilde
  \right) \nonumber\\
  &  -\med \vvec_{r}'^T\diag\left(\Exp{\alphavec}\right)\vvec_{r}'
  -\med \beta_{r}\left(\mu_r-\mu_{0_r}\right)^2
  +\const\\
  =&-\med\trace\left(-2\vtildevec'_{r}
    \left(\wbar_{rr}\Cmat_{r}+
      \sum_{s\neq r} \wbar_{rs}
      \left(\Cmat_{s} -\Exp{\vtildevec'_{s}}^{T}\Rmatytilde\right)\right)
    +\vtildevec'_r\vtildevec'^{T}_r\wbar_{rr}\Rmatytilde
  \right) \nonumber\\
  &  -\med \vvec_{r}'^T\diag\left(\Exp{\alphavec}\right)\vvec_{r}'
  -\med \beta_{r}\left(\mu_r-\mu_{0_r}\right)^2
  +\const\\
  =&-\med\trace\left(-2\vtildevec'_{r}
    \left(\wbar_{rr}\Cmat_{r}+
      \sum_{s\neq r} \wbar_{rs}
      \left(\Cmat_{s} -\Exp{\vtildevec'_{s}}^{T}\Rmatytilde\right)\right)
    +\vtildevec'_r\vtildevec'^{T}_r\wbar_{rr}\Rmatytilde
  \right) \nonumber\\
  &  -\med \vtildevec_{r}'^T\diag\left(\alphatbarvec_r\right)\vtildevec_{r}'
  +\beta_{r}\mu_r\mu_{0_r}
  +\const\\
  =&-\med\trace\left(-2\vtildevec'_{r}
    \left(\wbar_{rr}\Cmat_{r}+
      \sum_{s\neq r} \wbar_{rs}
      \left(\Cmat_{s} -\Exp{\vtildevec'_{s}}^{T}\Rmatytilde\right)
      +\beta_r\mutildevec_{0_r}^T\right)\right.\nonumber\\
  &\left.+\vtildevec'_r\vtildevec'^{T}_r
    \left(\diag\left(\alphatbarvec_r\right)+\wbar_{rr}\Rmatytilde\right)
  \right)
\end{align}
where $\wbar_{rs}$ is the element $r,s$ of $\Exp{\Wmat}$, 
\begin{align}
  \Cmat=&\sumiM\Fvec_i\Exp{\ytildevec_i}^{T}\\
  \Rmatytilde=&\sumiM N_i \Exp{\ytildevec_i\ytildevec_i^T}\\
  \Exp{\scatt{\ytildevec_i}}=&
  \begin{bmatrix}
    \Exp{\scatt{\yvec_i}} & \Exp{\yvec_i} \\
    \Exp{\yvec_i}^T & 1 
  \end{bmatrix}\\
  \alphatbarvec_r=&
  \begin{bmatrix}
    \Exp{\alphavec}\\
    \beta_r
  \end{bmatrix}
  \quad\quad 
  \mutildevec_{0_r}=
  \begin{bmatrix}
    \zerovec_{n_y \times 1}\\
    \mu_{0_r}
  \end{bmatrix}
\end{align}
and $\Cmat_r$ is the $r^{th}$ row of $\Cmat$.

Then $\qopt{\vtildevec'_r}$ is a Gaussian distribution:
\begin{align}
  \qopt{\vtildevec'_r}=&
  \Gauss{\vtildevec_{r}'}{\vtbarvec_{r}'}{\iLmatVtr}\\
  \LmatVtr=&\diag\left(\alphatbarvec_r\right)+\wbar_{rr}\Rmatytilde\\
  \vtbarvec_{r}'=&\iLmatVtr\left(\wbar_{rr}\Cmat_{r}^T+
    \sum_{s\neq r} \wbar_{rs}
    \left(\Cmat_{s}^T -\Rmatytilde\vtbarvec_{s}'\right)
    +\beta_r\mutildevec_{0_r}\right)
\end{align}

The optimum for $\qopt{\alphavec}$:
\begin{align}
  \lnqopt{\alphavec}=&
  \Expcond{\lnProb{\Phimat,\Ymat,\muvec,\Vmat,\Wmat,\alphavec}}
  {\Ymat,\muvec,\Vmat,\Wmat}+\const\\
  =&\Expcond{\lnProb{\Vmat|\alphavec}}{\Vmat}+\lnProb{\alphavec|a_\alpha,b_\alpha}+\const\\
  =& \sum_{q=1}^{n_y} \frac{d}{2}\ln \alpha_q -\med\alpha_q\Exp{\vvec_q^T\vvec_q}
  +(a_{\alpha}-1)\ln \alpha_q-b_{\alpha}\alpha_q+\const\\
  =& \sum_{q=1}^{n_y} \left(\frac{d}{2}+a_\alpha-1\right)\ln \alpha_q 
  -\alpha_q\left(b_{\alpha}+\med\Exp{\vvec_q^T\vvec_q}\right) +\const\\
\end{align}
Then $\qopt{\alphavec}$ is a product of Gammas:
\begin{align}
  \label{eq:baysplda_v1_apost}
  \qopt{\alphavec}=&\prod_{q=1}^{n_y}
  \Gammad{\alpha_q}{a'_{\alpha}}{b_{\alpha_q}'}\\
  a_{\alpha}'=&a_{\alpha}+\frac{d}{2}\\
  b_{\alpha_q}'=&b_{\alpha}+\med\Exp{\vvec_q^T\vvec_q}
\end{align}

The optimum for $\qopt{\Wmat}$ in the non-informative case:
\begin{align}
  \lnqopt{\Wmat}=&
  \Expcond{\lnProb{\Phimat,\Ymat,\muvec,\Vmat,\Wmat,\alphavec}}
  {\Ymat,\muvec,\Vmat,\alphavec}+\const\\
  =&\Expcond{\lnProb{\Phimat|\Ymat,\muvec,\Vmat,\Wmat}}{\Ymat,\muvec,\Vmat}
  +\lnProb{\Wmat}+\const\\
  =&\frac{N}{2}\lndet{\Wmat}-\frac{d+1}{2}\lndet{\Wmat}
  -\med\trace\left(\Wmat\Kmat\right)+\const
\end{align}
where
\begin{align}
  \Kmat=&\sumiM \Exp{\Smat_i
    -\Fvec_i\ytildevec_i^{T}\Vtildemat^{T}-\Vtildemat\ytildevec_i\Fvec_i^T
    +N_i \Vtildemat\ytildevec_i\ytildevec_i^T\Vtildemat^T}\\
  =&\Smat-\Cmat\Exp{\Vtildemat}^T-\Exp{\Vtildemat}\Cmat^T
  +\Expcond{\Vtildemat\Rmatytilde\Vtildemat^T}{\Vtildemat}
\end{align}
Then $\qopt{\Wmat}$ is Wishart distributed:
\begin{align}
  \Prob{\Wmat}=&\Wishart{\Wmat}{\Psimat}{\nu} \quad \textrm{if $\nu>d$}\\
  \iPsimat=&\Kmat\\
  \nu=&N
\end{align}

The optimum for $\qopt{\Wmat}$ in the informative case:
\begin{align}
  \lnqopt{\Wmat}=&
  \Expcond{\lnProb{\Phimat,\Ymat,\muvec,\Vmat,\Wmat,\alphavec}}
  {\Ymat,\muvec,\Vmat,\alphavec}+\const\\
  =&\Expcond{\lnProb{\Phimat|\Ymat,\muvec,\Vmat,\Wmat}}{\Ymat,\muvec,\Vmat}
  +\lnProb{\Wmat}+\const\\
  =&\frac{N}{2}\lndet{\Wmat}+\frac{\nud-d-1}{2}\lndet{\Wmat}
  -\med\trace\left(\Wmat\left(\iPsimat_0+\Kmat\right)\right)+\const
\end{align}
Then $\qopt{\Wmat}$ is Wishart distributed:
\begin{align}
  \Prob{\Wmat}=&\Wishart{\Wmat}{\Psimat}{\nu}\\
  \iPsimat=&\iPsimat_0+\Kmat\\
  \nu=&\nud+N
\end{align}

Finally, we evaluate the expectations:
\begin{align}
  \Exp{\alpha_q}=&\frac{a_{\alpha}^\prime}{b_{\alpha_q}^\prime}\\
  \Vtbarmat=&\Exp{\Vtildemat}=
  \begin{bmatrix}
    \vtbarvec_{1}'^T\\
    \vtbarvec_{2}'^T\\
    \vdots\\
    \vtbarvec_{d}'^T
  \end{bmatrix}\\
  \Wbarmat=&\Exp{\Wmat}=\nu\Psimat\\
  \Exp{\vvec_q^T\vvec_q}=&\sumrd\Exp{\vvec_{rq}'^T\vvec_{rq}'}\\
  =&\sumrd\iLmat_{\Vtildemat_{r qq}}+\vbarvec_{rq}'^2\\
  \Exp{\Vmat^T\Wmat\Vmat}=&\Exp{\Vmat'\Wbarmat\Vmat'^T}\\
  =&\sumrd\sumsd \wbar_{rs} \Exp{\vvec'_r\vvec'^T_s} \\
  =&\sumrd \wbar_{rr} \SigmatVr 
  +\sumrd\sumsd \wbar_{rs} \vbarvec'_r\vbarvec'^T_s\\
  =&\sumrd \wbar_{rr} \SigmatVr + \Vbarmat^T\Wbarmat\Vbarmat\\
  \Exp{\Vmat^T\Wmat\muvec}=&\sumrd \wbar_{rr} \SigmatVmur
  +\sumrd\sumsd\wbar_{rs}\vbarvec'_r\overline{\mu}_s\\
  =&\sumrd \wbar_{rr} \SigmatVmur+\Vbarmat^T\Wbarmat\mubarvec\\
  \Exp{\Vtildemat\Rmatytilde\Vtildemat^T}=&
  \sum_{r=1}^{n_y}\sum_{s=1}^{n_y}
  \rytilders\Exp{\vtildevec_r\vtildevec_s^T}\\
  =&\sum_{r=1}^{n_y}\sum_{s=1}^{n_y}
  \rytilders 
  \begin{bmatrix}
    \Exp{\vtilde_{r_i}\vtilde_{s_j}}
  \end{bmatrix}_{d\times d}\\
  =&\sum_{r=1}^{n_y}\sum_{s=1}^{n_y}
  \rytilders 
  \begin{bmatrix}
    \Exp{\vtilde'_{i_r}\vtilde'_{j_s}}
  \end{bmatrix}_{d\times d}\\
  =&\sum_{r=1}^{n_y}\sum_{s=1}^{n_y}
  \rytilders 
  \begin{bmatrix}
    \Exp{\vtilde'_{i_r}}\Exp{\vtilde'_{j_s}}
  \end{bmatrix}_{d\times d}
  +\sum_{r=1}^{n_y}\sum_{s=1}^{n_y}
  \rytilders \diag\left(
    \begin{bmatrix}
      \sigma_{\Vtildemat_{i_{rs}}}\\
    \end{bmatrix}_{d}\right)\\
  =&\sum_{r=1}^{n_y}\sum_{s=1}^{n_y}
  \rytilders 
  \begin{bmatrix}
    \Exp{\vtilde_{r_i}}\Exp{\vtilde_{s_j}}
  \end{bmatrix}_{d\times d}
  +\diag\left(\rhovec\right)\\
  =&\sum_{r=1}^{n_y}\sum_{s=1}^{n_y}
  \rytilders \Exp{\vtildevec_r}\Exp{\vtildevec_s}^T
  +\diag\left(\rhovec\right)\\
  =&\Vtbarmat\Rmatytilde\Vtbarmat^T+\diag\left(\rhovec\right)
\end{align}
where $\vtilde_{r_i}$ is the $i^{th}$ element of $\vtildevec_{r}$, 
$\vtilde'_{i_r}$ is the $r^{th}$ element of $\vtildevec'_{i}$,
\begin{align}
  \SigmatVtr=&
  \begin{bmatrix}
    \SigmatVr & \SigmatVmur\\
    \SigmatVmur^T & \Sigmatmur
  \end{bmatrix}
  =
  \begin{bmatrix}
    \sigma_{\Vtildemat_{r_{ij}}}
  \end{bmatrix}_{n_y \times n_y}
  =\iLmatVtr\\
  \rhovec=&
  \begin{bmatrix}
    \rho_1 & \rho_2 & \hdots & \rho_d
  \end{bmatrix}^T\\
  \rho_i=&\sum_{r=1}^{n_y}\sum_{s=1}^{n_y}\left(\Rmatytilde\circ\iLmatVti\right)_{rs}
\end{align}
and $\circ$ is the Hadamard product. 

\subsubsection{Distributions with deterministic annealing}

If we use annealing, for a parameter $\kappa$, we have:

\begin{align}
  \qopt{\Ymat}=&\prodiM \Gauss{\yvec_i}{\ybarvec_i}{1/\kappa\;\iLmatyi}\\
  \qopt{\vtildevec'_r}=&
  \Gauss{\vtildevec_{r}'}{\vtbarvec_{r}'}{1/\kappa\;\iLmatVtr}\\
  \qopt{\Wmat}=&\Wishart{\Wmat}{1/\kappa\;\Psimat}{\kappa(\nu-d-1)+d+1}
  \quad \textrm{if $\kappa(\nu-d-1)+1>0$}\\
  \qopt{\alphavec}=&\prod_{q=1}^{n_y}
  \Gammad{\alpha_q}{a'_{\alpha}}{b_{\alpha_q}'}\\
  a_{\alpha}'=&\kappa\left(a_{\alpha}+\frac{d}{2}-1\right)+1\\
  b_{\alpha_q}'=&\kappa\left(b_{\alpha}+\med\Exp{\vvec_q^T\vvec_q}\right)\;.
\end{align}

\subsection{Variational lower bound}
\label{sec:baysplda_v1_lb}

The lower bound is given by
\begin{align}
  \lowb=&\Expcond{\lnProb{\Phimat|\Ymat,\muvec,\Vmat,\Wmat}}
  {\Ymat,\muvec,\Vmat,\Wmat}
  +\Expcond{\lnProb{\Ymat}}{\Ymat}
  +\Expcond{\lnProb{\Vmat|\alphavec}}{\Vmat,\alphavec}
  \nonumber\\
  &+\Expcond{\lnProb{\alphavec}}{\alphavec}
  +\Expcond{\lnProb{\muvec}}{\muvec}
  +\Expcond{\lnProb{\Wmat}}{\Wmat}
  \nonumber\\
  &-\Expcond{\lnq{\Ymat}}{\Ymat}
  -\Expcond{\lnq{\Vtildemat}}{\Vtildemat}
  -\Expcond{\lnq{\alphavec}}{\alphavec}
  -\Expcond{\lnq{\Wmat}}{\Wmat}
\end{align}

The term $\Expcond{\lnProb{\Phimat|\Ymat,\muvec,\Vmat,\Wmat}}
{\Ymat,\muvec,\Vmat,\Wmat}$:
\begin{align}
  \Expcond{\lnProb{\Phimat|\Ymat,\muvec,\Vmat,\Wmat}}
  {\Ymat,\muvec,\Vmat,\Wmat}=&
  \frac{N}{2}\Exp{\lndet{\Wmat}}-\frac{Nd}{2}\ln(2\pi) \nonumber\\
  &-\med\trace\left(\Wbarmat\left(
      \Smat-2\Cmat\Vtbarmat^T+\Exp{\Vtildemat\Rmatytilde\Vtildemat^T}
    \right)\right)\\
  =&\frac{N}{2}\ln\overline{\Wmat}-\frac{Nd}{2}\ln(2\pi)
  -\med\trace\left(\Wbarmat\Smat\right) \nonumber\\
  &-\med\trace\left(-2\Vtbarmat^T\Wbarmat\Cmat
    +\Exp{\Vtildemat^T\Wmat\Vtildemat}\Rmatytilde
  \right)
\end{align}
where
\begin{align}
  \ln\overline{\Wmat}=&\Exp{\lndet{\Wmat}}\\
  =&\sumid\psi\left(\frac{\nu+1-i}{2}\right)+d\ln2 +\lndet{\Psimat}
\end{align}
and $\psi$ is the digamma function.

The term $\Expcond{\lnProb{\Ymat}}{\Ymat}$:
\begin{align}
  \Expcond{\lnProb{\Ymat}}{\Ymat}=&
  -\frac{M n_y}{2}\ln(2\pi)-\med\trace\left(\sumiM
    \Exp{\scatt{\yvec_i}}\right)\\
  =&-\frac{M n_y}{2}\ln(2\pi)-\med\trace\left(\Rhomat\right)
\end{align}
where
\begin{align}
  \Rhomat=\sumiM \Exp{\scatt{\yvec_i}}
\end{align}

The term $\Expcond{\lnProb{\Vmat|\alphavec}}{\Vmat,\alphavec}$:
\begin{align}
  \Expcond{\lnProb{\Vmat|\alphavec}}{\Vmat,\alphavec}=&
  -\frac{n_y d}{2}\ln(2\pi)+\frac{d}{2}\sum_{q=1}^{n_y}\Exp{\ln\alpha_q}
  -\med\sum_{q=1}^{n_y} \Exp{\alpha_q}\Exp{\vvec_q^T\vvec_q}
\end{align}
where
\begin{align}
  \Exp{\ln\alpha_q}=\psi(a_\alpha^\prime)-\ln b_{\alpha_q}^\prime \;.
\end{align}

The term $\Expcond{\lnProb{\alphavec}}{\alphavec}$:
\begin{align}
  \Expcond{\lnProb{\alphavec}}{\alphavec}=&
  n_y \left(a_\alpha\ln b_\alpha-\ln\gammaf{a_\alpha}\right)+
  \sum_{q=1}^{n_y}(a_\alpha-1)\Exp{\ln\alpha_q}-b_{\alpha}\Exp{\alpha_q}\\
  =& n_y \left(a_\alpha\ln b_\alpha-\ln\gammaf{a_\alpha}\right)+
  (a_\alpha-1)\sum_{q=1}^{n_y}\Exp{\ln\alpha_q}-b_{\alpha}\sum_{q=1}^{n_y}\Exp{\alpha_q}
\end{align}

The term $\Expcond{\lnProb{\muvec}}{\muvec}$:
\begin{align}
  \Expcond{\lnProb{\muvec}}{\muvec}=&
  -\frac{d}{2}\ln(2\pi)+\frac{1}{2}\sumrd\ln\beta_r
  -\med\sumrd\beta_r\left(\Exp{\mu_r^2}
    -2\mu_{0_r}\Exp{\mu_r}+\mu_{0_r}^2\right)\\
  =&-\frac{d}{2}\ln(2\pi)+\frac{1}{2}\sumrd\ln\beta_r
  -\med\sumrd\beta_r\left(\Sigmatmur+\Exp{\mu_r}^2
    -2\mu_{0_r}\Exp{\mu_r}+\mu_{0_r}^2\right)
\end{align}

The term $\Expcond{\lnProb{\Wmat}}{\Wmat}$ for the non-informative case:
\begin{align}
  \Expcond{\lnProb{\Wmat}}{\Wmat}=&-\frac{d+1}{2}\ln\overline{\Wmat}\;.
\end{align}

The term $\Expcond{\lnProb{\Wmat}}{\Wmat}$ for the informative case:
\begin{align}
  \Expcond{\lnProb{\Wmat}}{\Wmat}=&\ln B\left(\Psimat_0,\nud\right)
  +\frac{\nud-d-1}{2}\ln\overline{\Wmat}
  -\frac{\nu}{2}\trace\left(\iPsimat_0\Psimat\right)
\end{align}

The term $\Expcond{\lnq{\Ymat}}{\Ymat}$:
\begin{align}
  \Expcond{\lnq{\Ymat}}{\Ymat}=&
  -\frac{Mn_y}{2}\ln(2\pi)+\med\sumiM\lndet{\Lmatyi}
  -\med\trace\left(\Lmatyi\Exp{\scattp{\yvec_i-\ybarvec_i}}\right)\\
  =&-\frac{Mn_y}{2}\ln(2\pi)+\med\sumiM\lndet{\Lmatyi} \nonumber\\
  &-\med\sumiM\trace\left(\Lmatyi\left(
      \Exp{\scatt{\yvec_i}}
      -\ybarvec_i\Exp{\yvec_i}^T-\Exp{\yvec_i}\ybarvec_i^T
      +\scatt{\ybarvec_i}\right)\right)\\
  =&-\frac{Mn_y}{2}\ln(2\pi)+\med\sumiM\lndet{\Lmatyi}
  -\med\sumiM\trace\left(\Imat\right)\\
  =&-\frac{Mn_y}{2}(\ln(2\pi)+1)+\med\sumiM\lndet{\Lmatyi}
\end{align}

The term $\Expcond{\lnq{\Vtildemat}}{\Vtildemat}$:
\begin{align}
  \Expcond{\lnq{\Vtildemat}}{\Vtildemat}=&
  -\frac{d(n_y+1)}{2}\left(\ln(2\pi)+1\right)
  +\med\sumrd\lndet{\LmatVtr}
\end{align}

The term $\Expcond{\lnq{\alphavec}}{\alphavec}$:
\begin{align}
  \Expcond{\lnq{\alphavec}}{\alphavec}=&
  -\sum_{q=1}^{n_y}\Entrop{\q{\alpha_q}}\\
  =&\sum_{q=1}^{n_y}(a_\alpha^\prime-1)\psi(a_\alpha^\prime)
  +\ln b_{\alpha_q}^\prime-a_\alpha^\prime-\ln\gammaf{a_\alpha^\prime}\\
  =&n_y\left((a_\alpha^\prime-1)\psi(a_\alpha^\prime)
    -a_\alpha^\prime-\ln\gammaf{a_\alpha^\prime}\right)
  +\sum_{q=1}^{n_y}\ln b_{\alpha_q}^\prime
\end{align}

The term $\Expcond{\lnq{\Wmat}}{\Wmat}$:
\begin{align}
  \label{eq:EntropWishartW}
  \Expcond{\lnq{\Wmat}}{\Wmat}=&-\Entrop{\q{\Wmat}}\\
  =&\ln B\left(\Psimat,\nu\right)+\frac{\nu-d-1}{2}\ln\overline{\Wmat}
  -\frac{\nu d}{2}
\end{align}
where
\begin{align}
  B(\Amat,N)&=\frac{1}{2^{Nd/2}\WZNd}\left|\Amat\right|^{-N/2}\\
  \WZNd=&\WishartZNd{N}{d}
\end{align}

\subsection{Hyperparameter optimization}
\label{sec:baysplda_v1_hyp}

We can set the Hyperparameters
$\left(\muvec_0,\betavec,a_{\alpha},b_{\alpha}\right)$ 
manually or estimate them from the
development data maximizing the lower bound.

we derive for $a_{\alpha}$
\begin{align}
  \frac{\partial\lowb}{\partial a_{\alpha}}=&
  n_y\left(\ln b_{\alpha}-\psi(a_{\alpha})\right)
  +\sum_{q=1}^{n_y}\Exp{\ln\alpha_q}=0 \quad \implies\\
  \psi(a_{\alpha})=&\ln b_{\alpha}+\frac{1}{n_y}\sum_{q=1}^{n_y}\Exp{\ln\alpha_q}
\end{align}

We derive for $b_{\alpha}$:
\begin{align}
  \frac{\partial\lowb}{\partial b_{\alpha}}=&
  \frac{n_y a_{\alpha}}{b_\alpha}-\sum_{q=1}^{n_y}\Exp{\alpha_q}=\zerovec\quad \implies\\
  b_{\alpha}=&\left( \frac{1}{n_y a_{\alpha}}\sum_{q=1}^{n_y}\Exp{\alpha_q}\right)^{-1}
\end{align}

We solve these equations with the procedure described
in~\cite{Beal2003}. We write
\begin{align}
  \psi(a)=&\ln b+c\\
  b=&\frac{a}{d}
\end{align}
where
\begin{align}
  c=&\frac{1}{n_y}\sum_{q=1}^{n_y}\Exp{\ln\alpha_q}\\
  d=&\frac{1}{n_y}\sum_{q=1}^{n_y}\Exp{\alpha_q}
\end{align}
Then
\begin{align}
  f(a)=\psi(a)-\ln a + \ln d -c=0
\end{align}

We can solve for $a$ using Newton-Rhaphson iterations:
\begin{align}
  a_{new}=&a-\frac{f(a)}{f^\prime(a)}=\\
  =&a\left(1-\frac{\psi(a)-\ln a + \ln d -c}{a\psi^\prime(a)-1}\right)
\end{align}
This algorithm does not assure that $a$ remains positive. We can put a
minimum value for $a$. Alternatively we can solve the equation for
$\tilde{a}$ such as $a=exp(\tilde{a})$.
\begin{align}
  \tilde{a}_{new}=&\tilde{a}-\frac{f(\tilde{a})}{f^\prime(\tilde{a})}=\\
  =&\tilde{a}-\frac{\psi(a)-\ln a + \ln d -c}{\psi^\prime(a)a-1}
\end{align}
Taking exponential in both sides:
\begin{align}
  a_{new}=a\exp\left(-\frac{\psi(a)-\ln a + \ln d -c}{\psi^\prime(a)a-1}\right)
\end{align}

We derive for $\muvec_0$:
\begin{align}
  \frac{\partial\lowb}{\partial \muvec_0}=&\zerovec \quad \implies\\
  \muvec_0=&\Exp{\muvec}
\end{align}

We derive for $\betavec$:
\begin{align}
  \frac{\partial\lowb}{\partial \betavec}=&\zerovec \quad \implies\\
  \betavec_r^{-1}=&\Sigmatmur+\Exp{\mu_r}^2
  -2\mu_{0_r}\Exp{\mu_r}+\mu_{0_r}^2
\end{align}

If we take an isotropic prior for $\muvec$:
\begin{align}
  \betavec^{-1}=&\frac{1}{d}\sumrd \Sigmatmur+\Exp{\mu_r}^2
  -2\mu_{0_r}\Exp{\mu_r}+\mu_{0_r}^2
\end{align}

\subsection{Minimum divergence}
\label{sec:baysplda_v1_md}

We assume a more general prior for the hidden variables:
\begin{align}
  \Prob{\yvec}=\Gauss{\yvec}{\muvecy}{\iLambmaty}
\end{align}
To minimize the divergence we maximize the part of $\lowb$ that depends
on $\muvecy$:
\begin{align}
  \lowb(\muvecy,\Lambmaty)=&\sumiM \ExpcondY{\ln
    \Gauss{\yvec}{\muvecy}{\iLambmaty}}
\end{align}

The, we get
\begin{align}
  \muvecy=&\frac{1}{M}\sumiM\ExpcondY{\yvec_i}\\
  \Sigmaty=&\Lambmaty^{-1}
  =\frac{1}{M}\sumiM\ExpcondY{\scattp{\yvec_i-\muvecy}}\\
  =&\frac{1}{M}\sumiM\ExpcondY{\scatt{\yvec_i}}-\scatt{\muvecy}
\end{align}

We have a transform $\yvec=\phi(\yvec^\prime)$ such as
$\yvec^\prime$ has a standard prior:
\begin{align}
  \yvec=&\muvecy+(\Sigmaty^{1/2})^{T}\yvec^\prime
\end{align}

we also can write that as
\begin{align}
  \ytildevec=&\Jmat\ytildevec^\prime
\end{align}
where
\begin{align}
  \Jmat=
  \begin{bmatrix}
    (\Sigmaty^{1/2})^T & \muvecy \\
    \zerovec^T & 1
  \end{bmatrix}
\end{align}
Now, we get $\q{\vtildevec'_r}$ such us if we apply the transform
$\yvec'=\phivec^{-1}(\yvec)$, the term 
$\Exp{\lnProb{\Xmat|\Ymat,\Wmat}}$ of 
$\lowb$ remains constant:
\begin{align}
  \vtbarvec_{r}'\leftarrow&\Jmat^T\vtbarvec_{r}'\\
  \iLmatVtr\leftarrow&\Jmat^T\iLmatVtr\Jmat\\
  \LmatVtr\leftarrow&\Gmat^T\LmatVtr\Gmat
\end{align}
where
\begin{align}
  \Gmat=&\left(\Jmat^T\right)^{-1}\\
  =&
  \begin{bmatrix}
    (\Sigmaty^{1/2})^{-1}
    & \zerovec\\
    -\muvecy^T(\Sigmaty^{1/2})^{-1} & 1
  \end{bmatrix}
\end{align}

\section{Variational inference with Gaussian-Gamma priors for $\Vmat$,
  Gaussian for $\muvec$ and Gamma for $\Wmat$}
\label{sec:baysplda_v2}

\subsection{Model priors}
\label{sec:baysplda_v2_priors}

In section~\ref{sec:baysplda_v1}, 
we saw that if we use a full covariance $\Wmat$ we had a full
covariance posterior for $\Vtildemat$. Then, to get a tractable
solution, we forced independence between the the rows of $\Vtildemat$
when choosing the variational partition function. 

In this section, we are going to assume that we have applied a
rotation to the data such as we can consider that $\Wmat$ is going to
remain diagonal during the VB iteration.

Then we are going to place a broad Gamma prior over each element of the
diagonal of $\Wmat$:
\begin{align}
  \Prob{\Wmat}=&\prodrd\Gammad{w_{rr}}{a_{w}}{b_{w}}
\end{align} 

We also consider the case of an isotropic $\Wmat$
($\Wmat=w\Imat$). Then the prior is
\begin{align}
  \Prob{w}=&\Gammad{w}{a_{w}}{b_{w}}
\end{align}

\subsection{Variational distributions}
\label{sec:baysplda_v2_q}

We write the joint distribution of the latent variables:
\begin{align}
  \Prob{\Phimat,\Ymat,\muvec,\Vmat,\Wmat,\alphavec
    |\muvec_0,\betavec,a_{\alpha},b_{\alpha},a_{w},b_{w}}=&
  \Prob{\Phimat|\Ymat,\muvec,\Vmat,\Wmat}\Prob{\Ymat}
  \Prob{\Vmat|\alphavec}\nonumber\\
  &\Prob{\alphavec|a,b}
  \Prob{\muvec|\muvec_0,\beta}\Prob{\Wmat|a_{w},b_{w}}
\end{align}
Following, the conditioning on
$\muvec_0,\betavec,a_{\alpha},b_{\alpha},a_{w},b_{w}$ 
will be dropped for convenience. 

Now, we consider the partition of the posterior:
\begin{align}
  \Prob{\Ymat,\muvec,\Vmat,\Wmat,\alphavec|\Phimat}\approx
  \q{\Ymat,\muvec,\Vmat,\Wmat,\alphavec}=
  \q{\Ymat}\q{\Vtildemat}\q{\Wmat}\q{\alphavec}
\end{align}

The optimum for $\qopt{\Ymat}$ and $\qopt{\alphavec}$ are the same as in 
section~\ref{sec:baysplda_v1_q}.

The optimum for $\qopt{\Vtildemat}$:
\begin{align}
  \lnqopt{\Vtildemat}=&
  \Expcond{\lnProb{\Phimat,\Ymat,\muvec,\Vmat,\Wmat,\alphavec}}
  {\Ymat,\Wmat,\alphavec}+\const\\
  =&\Expcond{\lnProb{\Phimat|\Ymat,\muvec,\Vmat,\Wmat}}{\Ymat,\Wmat}
  +\Expcond{\lnProb{\Vmat|\alphavec}}{\alphavec}+\lnProb{\muvec}+\const\\
  =&-\med\trace\left(-2\Vtildemat^{T}\Exp{\Wmat}\Cmat
    +\Vtildemat^{T}\Exp{\Wmat}\Vtildemat\Rmatytilde
  \right) \nonumber\\
  &-\med \sumrd \vvec_{r}'^T\diag\left(\Exp{\alphavec}\right)\vvec_{r}'
  -\med \sum_{r=1}^{d} \beta_{r}\left(\mu_r-\mu_{0_r}\right)^2
  +\const\\
  =&-\med\sumrd\trace\left(-2\vtildevec'_{r}
    \left(\wbar_{rr}\Cmat_{r}
      +\beta_r\mutildevec_{0_r}^T\right)
    +\vtildevec'_r\vtildevec'^{T}_r
    \left(\diag\left(\alphatbarvec_r\right)+\wbar_{rr}\Rmatytilde\right)
  \right)
\end{align}

Then $\qopt{\Vtildemat}$ is a product of Gaussian distributions:
\begin{align}
  \qopt{\Vtildemat}=&\prodrd
  \Gauss{\vtildevec_{r}'}{\wtbarvec_{r}'}{\iLmatVtr}\\
  \LmatVtr=&\diag\left(\alphatbarvec_r\right)+\wbar_{rr}\Rmatytilde\\
  \vtbarvec_{r}'=&\iLmatVtr\left(\wbar_{rr}\Cmat_{r}^T
    +\beta_r\mutildevec_{0_r}\right)
\end{align}

The optimum for $\qopt{\Wmat}$:
\begin{align}
  \lnqopt{\Wmat}=&
  \Expcond{\lnProb{\Phimat,\Ymat,\muvec,\Vmat,\Wmat,\alphavec}}
  {\Ymat,\muvec,\Vmat,\alphavec}+\const\\
  =&\Expcond{\lnProb{\Phimat|\Ymat,\muvec,\Vmat,\Wmat}}{\Ymat,\muvec,\Vmat}
  +\lnProb{\Wmat}+\const\\
  =&\sumrd \frac{N}{2}\ln w_{rr}-\med w_{rr}k_{rr}
  +(a_{w}-1)\ln w_{rr}-b_{w}w_{rr}+\const\\
  =&\sumrd \left(a_{w}+\frac{N}{2}-1\right)\ln w_{rr}
  -\left(b_w+\med k_{rr} \right)w_{rr} +\const
\end{align}
where
\begin{align}
  \Kmat=&\diag\left(\Smat-\Cmat\Exp{\Vtildemat}^T-\Exp{\Vtildemat}\Cmat^T
    +\Expcond{\Vtildemat\Rmatytilde\Vtildemat^T}{\Vtildemat}\right)
\end{align}

Then $\qopt{\Wmat}$ is a product of Gammas:
\begin{align}
  \label{eq:baysplda_v2_wpost}
  \qopt{\Wmat}=&\prodrd
  \Gammad{w_{rr}}{a'_{w}}{b_{w_r}'}\\
  a_{w}'=&a_{w}+\frac{N}{2}\\
  b_{w_r}'=&b_{w}+\med k_{rr}
\end{align}

If we force an isotropic $\Wmat$, the optimum $\qopt{\Wmat}$ is
\begin{align}
  \lnqopt{\Wmat}=&
  \frac{Nd}{2}\ln w-\med wk
  +(a_{w}-1)\ln w-b_{w}w+\const\\
  =&\left(a_{w}+\frac{Nd}{2}-1\right)\ln w
  -\left(b_w+\med k \right)w +\const
\end{align}
where
\begin{align}
  k=&\trace\left(\Smat-\Cmat\Exp{\Vtildemat}^T-\Exp{\Vtildemat}\Cmat^T
    +\Expcond{\Vtildemat\Rmatytilde\Vtildemat^T}{\Vtildemat}\right)\\
  =&\trace\left(\Smat-2\Cmat\Exp{\Vtildemat}^T\right)
  +\trace\left(\Exp{\Vtildemat^T\Vtildemat}\Rmatytilde\right)
\end{align}

Then $\qopt{\Wmat}$ is a Gamma distribution:
\begin{align}
  \label{eq:baysplda_v2_wpost}
  \qopt{\Wmat}=&
  \Gammad{w}{a'_{w}}{b_w'}\\
  a_{w}'=&a_{w}+\frac{Nd}{2}\\
  b_{w}'=&b_{w}+\med k
\end{align}

Finally, we evaluate the expectations:
\begin{align}
  \Exp{\Vtildemat^T\Vtildemat}=&\Exp{\Vtildemat'\Vtildemat'^T}\\
  =&\sumrd \iLmatVtr+\Exp{\Vtildemat'}\Exp{\Vtildemat'}^T\\
  =&\sumrd \iLmatVtr+\Exp{\Vtildemat}^T\Exp{\Vtildemat}
\end{align}

\subsection{Variational lower bound}
\label{sec:baysplda_v2_lb}

The lower bound is given by
\begin{align}
  \lowb=&\Expcond{\lnProb{\Phimat|\Ymat,\muvec,\Vmat,\Wmat}}
  {\Ymat,\muvec,\Vmat,\Wmat}
  +\Expcond{\lnProb{\Ymat}}{\Ymat}
  +\Expcond{\lnProb{\Vmat|\alphavec}}{\Vmat,\alphavec}
  \nonumber\\
  &+\Expcond{\lnProb{\alphavec}}{\alphavec}
  +\Expcond{\lnProb{\muvec}}{\muvec}
  +\Expcond{\lnProb{\Wmat}}{\Wmat}
  \nonumber\\
  &-\Expcond{\lnq{\Ymat}}{\Ymat}
  -\Expcond{\lnq{\Vtildemat}}{\Vtildemat}
  -\Expcond{\lnq{\alphavec}}{\alphavec}
  -\Expcond{\lnq{\Wmat}}{\Wmat}
\end{align}

The term $\Expcond{\lnProb{\Phimat|\Ymat,\muvec,\Vmat,\Wmat}}
{\Ymat,\muvec,\Vmat,\Wmat}$:
\begin{align}
  \Expcond{\lnProb{\Phimat|\Ymat,\muvec,\Vmat,\Wmat}}
  {\Ymat,\muvec,\Vmat,\Wmat}=&
  \frac{N}{2}\sumrd\ln\overline{w}_{rr}-\frac{Nd}{2}\ln(2\pi)
  -\med\trace\left(\Wbarmat\Smat\right) \nonumber\\
  &-\med\trace\left(-2\Vtbarmat^T\Wbarmat\Cmat
    +\Exp{\Vtildemat^T\Wmat\Vtildemat}\Rmatytilde
  \right)
\end{align}
where
\begin{align}
  \ln\overline{w}_{rr}=&\Exp{\lndet{w_{rr}}}\\
  =&\psi(a_{w}')-\ln b_{w_r}'
\end{align}
and $\psi$ is the digamma function.

The term $\Expcond{\lnProb{\Wmat}}{\Wmat}$ with non-isotropic $\Wmat$:
\begin{align}
  \Expcond{\lnProb{\Wmat}}{\Wmat}=&
  \sumrd a_{w}\ln b_{w}+\left(a_{w}-1\right)\Exp{\ln w_{rr}}
  -b_{w}\Exp{w_{rr}}-\ln\gammaf{a_w}\\
  =&d\left(a_{w}\ln b_{w}-\ln\gammaf{a_w}\right)
  +\left(a_{w}-1\right)\sumrd\Exp{\ln w_{rr}}-b_w\sumrd\Exp{w_{rr}}
\end{align}

The term $\Expcond{\lnProb{\Wmat}}{\Wmat}$ with isotropic $\Wmat$:
\begin{align}
  \Expcond{\lnProb{\Wmat}}{\Wmat}=&
  a_{w}\ln b_{w}-\ln\gammaf{a_w}
  +\left(a_{w}-1\right)\Exp{\ln w}-b_w\Exp{w}
\end{align}

The term $\Expcond{\lnq{\Wmat}}{\Wmat}$ with non-isotropic $\Wmat$:
\begin{align}
  \Expcond{\lnq{\Wmat}}{\Wmat}=&-\sumrd\Entrop{\q{w_{rr}}}\\
  =&d\left((a_w'-1)\psi(a_w')
    -a_w'-\ln\gammaf{a_w'}\right)
  +\sumrd\ln b_{w_r}'
\end{align}

The term $\Expcond{\lnq{\Wmat}}{\Wmat}$ with isotropic $\Wmat$:
\begin{align}
  \Expcond{\lnq{\Wmat}}{\Wmat}=&-\Entrop{\q{w}}\\
  =&(a_w'-1)\psi(a_w')
  -a_w'-\ln\gammaf{a_w'}
  +\ln b_{w}'
\end{align}

The rest of terms are the same as in section~\ref{sec:baysplda_v2_lb}.

\subsection{Hyperparameter optimization}
\label{sec:baysplda_v2_hyp}

We can estimate the parameters $(a_{w},b_w)$ from the
development data maximizing the lower bound.

For non-isotropic $\Wmat$:

we derive for $a_{w}$
\begin{align}
  \frac{\partial\lowb}{\partial a_{w}}=&
  d\left(\ln b_{w}-\psi(a_{w})\right)
  +\sum_{r=1}^{d}\Exp{\ln w_{rr}}=0 \quad \implies\\
  \psi(a_{w})=&\ln b_{w}+\frac{1}{d}\sum_{r=1}^{d}\Exp{\ln w_{rr}}
\end{align}

We derive for $b_{w}$:
\begin{align}
  \frac{\partial\lowb}{\partial b_{w}}=&
  \frac{d a_{w}}{b}-\sum_{r=1}^{d}\Exp{w_{rr}}=\zerovec\quad \implies\\
  b_{w}=&\left( \frac{1}{d a_{w}}\sum_{r=1}^{d}\Exp{w_{rr}}\right)^{-1}
\end{align}

For isotropic $\Wmat$:

we derive for $a_{w}$
\begin{align}
  \frac{\partial\lowb}{\partial a_{w}}=&
  \ln b_{w}-\psi(a_{w})
  +\Exp{\ln w}=0 \quad \implies\\
  \psi(a_{w})=&\ln b_{w}+\Exp{\ln w}
\end{align}

We derive for $b_{w}$:
\begin{align}
  \frac{\partial\lowb}{\partial b_{w}}=&
  \frac{a_{w}}{b}-\Exp{w}=\zerovec\quad \implies\\
  b_{w}=&\left( \frac{1}{a_{w}}\Exp{w}\right)^{-1}
\end{align}

We can solve these equations by Newton-Rhapson iterations as described
in section~\ref{sec:baysplda_v1_hyp}.

\section{Variational inference with full covariance Gaussian prior for
  $\Vmat$ and $\muvec$ and Wishart for $\Wmat$}
\label{sec:baysplda_v3} 

\subsection{Model priors}
\label{sec:baysplda_v3_priors}

Lets assume that we compute
the posterior of model parameters given a development database with a large
amount of data. If we want to compute the model posterior for a
small database we could use the posterior given the large database as
prior. 

Thus, we take a prior distribution for $\Vtildemat$
\begin{align}
  \Prob{\Vtildemat}=\prodrd
  \Gauss{\vvec_{r}'}{\vbarvec_{0 r}'}{\iLmatVtdr}
\end{align}

The prior for $\Wmat$ is
\begin{align}
  \Prob{\Wmat}=\Wishart{\Wmat}{\Psimat_{0}}{\nud}
\end{align}
The parameters $\vbarvec_{0 r}'$, $\iLmatVtdr$, $\Psi_{0}$ and $\nud$ 
are computed with the large dataset.

\subsection{Variational distributions}
\label{sec:baysplda_v3_q}

The joint distribution of the latent variables:
\begin{align}
  \Prob{\Phimat,\Ymat,\muvec,\Vmat,\Wmat}=
  \Prob{\Phimat|\Ymat,\muvec,\Vmat,\Wmat}\Prob{\Ymat}
  \Prob{\Vtildemat}\Prob{\Wmat}
\end{align}

Now, we consider the partition of the posterior:
\begin{align}
  \Prob{\Ymat,\muvec,\Vmat,\Wmat|\Phimat}\approx
  \q{\Ymat,\Vtildemat,\Wmat}=
  \q{\Ymat}\prod_{r=1}^d\q{\vtildevec'_r}\q{\Wmat}
\end{align}

The optimum for $\qopt{\Ymat}$ is the same as in
section~\ref{sec:baysplda_v1_q}. 

The optimum for $\qopt{\vtildevec'_r}$:
\begin{align}
  \lnqopt{\vtildevec'_r}=&
  \Expcond{\lnProb{\Phimat,\Ymat,\muvec,\Vmat,\Wmat}}
  {\Ymat,\Wmat,\vtildevec'_{s\neq r}}+\const\\
  =&\Expcond{\lnProb{\Phimat|\Ymat,\muvec,\Vmat,\Wmat}}
  {\Ymat,\Wmat,\vtildevec'_{s\neq r}}
  +\lnProb{\Vtildemat}+\const\\
  =&-\med\trace\left(-2\vtildevec'_{r}
    \left(\wbar_{rr}\Cmat_{r}+
      \sum_{s\neq r} \wbar_{rs}
      \left(\Cmat_{s} -\Exp{\vtildevec'_{s}}^{T}\Rmatytilde\right)\right)
    +\vtildevec'_r\vtildevec'^{T}_r\wbar_{rr}\Rmatytilde
  \right) \nonumber\\
  &-\med \mahP{\vvec_{r}'}{\vbarvec_{0 r}'}{\LmatVtdr}
  +\const\\
  =&-\med\trace\left(-2\vtildevec'_{r}
    \left(\vbarvec_{0 r}'^T\LmatVtdr+\wbar_{rr}\Cmat_{r}+
      \sum_{s\neq r} \wbar_{rs}
      \left(\Cmat_{s} -\Exp{\vtildevec'_{s}}^{T}\Rmatytilde\right)
    \right)\right.\nonumber\\
  &\left.+\vtildevec'_r\vtildevec'^{T}_r
    \left(\LmatVtdr+\wbar_{rr}\Rmatytilde\right)
  \right)
\end{align}

Then $\qopt{\vtildevec'_r}$ is a Gaussian distribution:
\begin{align}
  \qopt{\vtildevec'_r}=&
  \Gauss{\vtildevec_{r}'}{\vtbarvec_{r}'}{\iLmatVtr}\\
  \LmatVtr=&\LmatVtdr+\wbar_{rr}\Rmatytilde\\
  \vtbarvec_{r}'=&\iLmatVtr\left(\LmatVtdr\vbarvec_{0 r}'
    +\wbar_{rr}\Cmat_{r}^T+\sum_{s\neq r} \wbar_{rs}
    \left(\Cmat_{s}^T -\Rmatytilde\vtbarvec_{s}'\right)
  \right)
\end{align}

The optimum for $\qopt{\Wmat}$:
\begin{align}
  \lnqopt{\Wmat}=&
  \Expcond{\lnProb{\Phimat,\Ymat,\muvec,\Vmat,\Wmat,\alphavec}}
  {\Ymat,\muvec,\Vmat,\alphavec}+\const\\
  =&\Expcond{\lnProb{\Phimat|\Ymat,\muvec,\Vmat,\Wmat}}{\Ymat,\muvec,\Vmat}
  +\lnProb{\Wmat}+\const\\
  =&\frac{N}{2}\lndet{\Wmat}+\frac{\nud-d-1}{2}\lndet{\Wmat}
  -\med\trace\left(\Wmat\left(\iPsimat_0+\Kmat\right)\right)+\const
\end{align}
where
\begin{align}
  \Kmat=&\Smat-\Cmat\Exp{\Vtildemat}^T-\Exp{\Vtildemat}\Cmat^T
  +\Expcond{\Vtildemat\Rmatytilde\Vtildemat^T}{\Vtildemat}
\end{align}
Then $\qopt{\Wmat}$ is Wishart distributed:
\begin{align}
  \Prob{\Wmat}=&\Wishart{\Wmat}{\Psimat}{\nu}\\
  \iPsimat=&\iPsimat_0+\Kmat\\
  \nu=&\nud+N
\end{align}

\subsubsection{Distributions with deterministic annealing}

If we use annealing, for a parameter $\kappa$, we have:

\begin{align}
  \qopt{\Wmat}=&\Wishart{\Wmat}{1/\kappa\;\Psimat}{\kappa(\nud+N-d-1)+d+1}
  \quad \textrm{if $\kappa(\nud+N-d-1)+1>0$}
\end{align}

\subsection{Variational lower bound}
\label{sec:baysplda_v3_lb}

The lower bound is given by
\begin{align}
  \lowb=&\Expcond{\lnProb{\Phimat|\Ymat,\muvec,\Vmat,\Wmat}}
  {\Ymat,\muvec,\Vmat,\Wmat}
  +\Expcond{\lnProb{\Ymat}}{\Ymat}
  +\Expcond{\lnProb{\Vtildemat}}{\Vtildemat}
  +\Expcond{\lnProb{\Wmat}}{\Wmat}
  \nonumber\\
  &-\Expcond{\lnq{\Ymat}}{\Ymat}
  -\Expcond{\lnq{\Vtildemat}}{\Vtildemat}
  -\Expcond{\lnq{\Wmat}}{\Wmat}
\end{align}

The term $\Expcond{\lnProb{\Vtildemat}}{\Vtildemat}$:
\begin{align}
  \Expcond{\lnProb{\Vtildemat}}{\Vtildemat}=&
  -\frac{n_y d}{2}\ln(2\pi)+\med\sumrd\lndet{\LmatVtdr} \nonumber\\
  &-\med\sumrd\trace\left(\LmatVtdr
    \Exp{\scattp{\vvec_{r}'-\vbarvec_{0 r}'}}\right)\\
  =&-\frac{n_y d}{2}\ln(2\pi)+\med\sumrd\lndet{\LmatVtdr} \nonumber\\
  &-\med \sumrd \trace\left(\LmatVtdr\left(\iLmatVtr
      +\scatt{\vbarvec_{r}'}
      -\vbarvec_{0 r}'\vbarvec_{r}'^T
      -\vbarvec_{r}'\vbarvec_{0 r}'^T
      +\scatt{\vbarvec_{0 r}'}\right)\right)\\
  =&-\frac{n_y d}{2}\ln(2\pi)+\med\sumrd\lndet{\LmatVtdr} \nonumber\\
  &-\med\sumrd  \trace\left(\LmatVtdr\iLmatVtr\right)
  -\med \sumrd 
  \mahP{\vbarvec_{r}'}{\vbarvec_{0 r}'}{\LmatVtdr}\\
  =&-\frac{n_y d}{2}\ln(2\pi)+\med\sumrd\lndet{\LmatVtdr} \nonumber\\
  &-\med\sumrd  \trace\left(\LmatVtdr\iLmatVtr\right)
  -\med\sumrd \trace\left(\LmatVtdr  
    \scattp{\vbarvec_{r}'-\vbarvec_{0 r}'}\right)
\end{align}

The term $\Expcond{\lnProb{\Wmat}}{\Wmat}$:
\begin{align}
  \Expcond{\lnProb{\Wmat}}{\Wmat}=&\ln B\left(\Psimat_0,\nud\right)
  +\frac{\nud-d-1}{2}\ln\overline{\Wmat}
  -\frac{\nu}{2}\trace\left(\iPsimat_0\Psimat\right)
\end{align}
where
\begin{align}
  \ln\overline{\Wmat}=&\Exp{\lndet{\Wmat}}\\
  =&\sumid\psi\left(\frac{\nu+1-i}{2}\right)+d\ln2 +\lndet{\Psimat}
\end{align}
and $\psi$ is the digamma function.

The term $\Expcond{\lnq{\Wmat}}{\Wmat}$
\begin{align}
  \Expcond{\lnq{\Wmat}}{\Wmat}=&-\Entrop{\q{\Wmat}}\\
  =&\ln B\left(\Psimat,\nu\right)+\frac{\nu-d-1}{2}\ln\overline{\Wmat}
  -\frac{\nu d}{2}
\end{align}

The rest of terms are the same as the ones in section~\ref{sec:baysplda_v1_lb}.

\section{Variational inference with full covariance Gaussian prior for
  $\Vmat$ and $\muvec$ and Gamma for $\Wmat$}
\label{sec:baysplda_v4} 

\subsection{Model priors}
\label{sec:baysplda_v4_priors}

Thus, we take a prior distribution for $\Vtildemat$
\begin{align}
  \Prob{\Vtildemat}=\prodrd
  \Gauss{\vvec_{r}'}{\vbarvec_{0 r}'}{\iLmatVtdr}
\end{align}

The prior for non-isotropic $\Wmat$ is
\begin{align}
  \Prob{\Wmat}=\prodrd
  \Gammad{w_{rr}}{a_{w}}{b_{w_r}}
\end{align}

The prior for isotropic $\Wmat$ is
\begin{align}
  \Prob{\Wmat}=
  \Gammad{w}{a_{w}}{b_{w}}
\end{align}

The parameters $\vbarvec_{0 r}'$, $\iLmatVtdr$, $a_w$ and $b_w$ 
are computed with the large dataset.

\subsection{Variational distributions}
\label{sec:baysplda_v4_q}

The joint distribution of the latent variables:
\begin{align}
  \Prob{\Phimat,\Ymat,\muvec,\Vmat,\Wmat}=
  \Prob{\Phimat|\Ymat,\muvec,\Vmat,\Wmat}\Prob{\Ymat}
  \Prob{\Vtildemat}\Prob{\Wmat}
\end{align}

Now, we consider the partition of the posterior:
\begin{align}
  \Prob{\Ymat,\muvec,\Vmat,\Wmat|\Phimat}\approx
  \q{\Ymat,\Vtildemat,\Wmat}=
  \q{\Ymat}\q{\Vtildemat}\q{\Wmat}
\end{align}

The optimum for $\qopt{\Ymat}$ is the same as in
section~\ref{sec:baysplda_v1_q}. 

Then optimum for $\qopt{\Vtildemat}$ is:
\begin{align}
  \qopt{\Vtildemat}=&\prodrd
  \Gauss{\vtildevec_{r}'}{\vtbarvec_{r}'}{\iLmatVtr}\\
  \LmatVtr=&\LmatVtdr+\wbar_{rr}\Rmatytilde\\
  \vtbarvec_{r}'=&\iLmatVtr\left(\LmatVtdr\vbarvec_{0 r}'
    +\wbar_{rr}\Cmat_{r}^T\right)
\end{align}

The optimum for $\qopt{\Wmat}$ for non-isotropic $\Wmat$:
\begin{align}
  \label{eq:baysplda_v2_wpost}
  \qopt{\Wmat}=&\prodrd
  \Gammad{w_{rr}}{a'_{w}}{b_{w_r}'}\\
  a_{w}'=&a_{w}+\frac{N}{2}\\
  b_{w_r}'=&b_{w_r}+\med k_{rr}
\end{align}
where
\begin{align}
  \Kmat=&\diag\left(\Smat-\Cmat\Exp{\Vtildemat}^T-\Exp{\Vtildemat}\Cmat^T
    +\Expcond{\Vtildemat\Rmatytilde\Vtildemat^T}{\Vtildemat}\right)
\end{align}

The optimum for $\qopt{\Wmat}$ for isotropic $\Wmat$:
\begin{align}
  \label{eq:baysplda_v2_wpost}
  \qopt{\Wmat}=&
  \Gammad{w}{a'_{w}}{b_w'}\\
  a_{w}'=&a_{w}+\frac{Nd}{2}\\
  b_{w}'=&b_{w}+\med k
\end{align}
where
\begin{align}
  k=&\trace\left(\Smat-2\Cmat\Exp{\Vtildemat}^T\right)
  +\trace\left(\Exp{\Vtildemat^T\Vtildemat}\Rmatytilde\right)
\end{align}

\subsection{Variational lower bound}
\label{sec:baysplda_v4_lb}

The lower bound is given by
\begin{align}
  \lowb=&\Expcond{\lnProb{\Phimat|\Ymat,\muvec,\Vmat,\Wmat}}
  {\Ymat,\muvec,\Vmat,\Wmat}
  +\Expcond{\lnProb{\Ymat}}{\Ymat}
  +\Expcond{\lnProb{\Vtildemat}}{\Vtildemat}
  +\Expcond{\lnProb{\Wmat}}{\Wmat}
  \nonumber\\
  &-\Expcond{\lnq{\Ymat}}{\Ymat}
  -\Expcond{\lnq{\Vtildemat}}{\Vtildemat}
  -\Expcond{\lnq{\Wmat}}{\Wmat}
\end{align}

The term $\Expcond{\lnProb{\Wmat}}{\Wmat}$ with non-isotropic $\Wmat$:
\begin{align}
  \Expcond{\lnProb{\Wmat}}{\Wmat}=&
  \sumrd a_{w}\ln b_{w_r}+\left(a_{w}-1\right)\Exp{\ln w_{rr}}
  -b_{w_r}\Exp{w_{rr}}-\ln\gammaf{a_w}\\
  =&-d\ln\gammaf{a_w}+ a_{w}\sumrd\ln b_{w_r}
  +\left(a_{w}-1\right)\sumrd\Exp{\ln w_{rr}}
  -\sumrd b_{w_r}\Exp{w_{rr}}
\end{align}

The terms $\Expcond{\lnProb{\Ymat}}{\Ymat}$,
$\Expcond{\lnq{\Ymat}}{\Ymat}$ and
$\Expcond{\lnq{\Vtildemat}}{\Vtildemat}$ 
are the same as in 
section~\ref{sec:baysplda_v1_lb}.

The terms $\Expcond{\lnProb{\Phimat|\Ymat,\muvec,\Vmat,\Wmat}}
{\Ymat,\muvec,\Vmat,\Wmat}$, $\Expcond{\lnProb{\Wmat}}{\Wmat}$ with
isotropic $\Wmat$ and $\Expcond{\lnq{\Wmat}}{\Wmat}$ 
are the same as is in
section~\ref{sec:baysplda_v2_lb}.

The term $\Expcond{\lnProb{\Vtildemat}}{\Vtildemat}$ is the same as in
section~\ref{sec:baysplda_v3_lb}.

\bibliographystyle{IEEEbib}
\bibliography{villalba}

\begin{thebibliography}{1}

\bibitem{villalba-splda}
Jes\'{u}s Villalba,
\newblock ``{SPLDA},''
\newblock Tech. {R}ep., University of Zaragoza, Zaragoza, 2011.

\bibitem{villalba-bay2cov-is2011}
Jes\'{u}s Villalba and Niko Brummer,
\newblock ``{Towards Fully Bayesian Speaker Recognition: Integrating Out the
  Between-Speaker Covariance},''
\newblock in {\em Interspeech 2011}, Florence, 2011, pp. 28--31.

\bibitem{Bishop1999}
C~M Bishop,
\newblock ``{Variational principal components},''
\newblock {\em 9th International Conference on Artificial Neural Networks ICANN
  99}, vol. 1, no. 470, pp. 509--514, 1999.

\bibitem{villalba-bay2cov}
Jes\'{u}s Villalba,
\newblock ``{Fully Bayesian Two-Covariance Model},''
\newblock Tech. {R}ep., University of Zaragoza, Zaragoza (Spain), 2010.

\bibitem{Beal2003}
Matthew~J Beal,
\newblock ``{Variational algorithms for approximate Bayesian inference},''
\newblock {\em Philosophy}, vol. 38, no. May, pp. 1--281, 2003.

\end{thebibliography}

\end{document}